\documentclass[a4paper]{jpconf}
\usepackage{graphicx}
\usepackage[english]{babel}

\begin{document}
\title{Creation of digital elevation models for river floodplains}

\author{Anna Klikunova$^{1*}$, Alexander Khoperskov$^{1}$}

\address{$^{1}$Volgograd State University, Volgograd, 400062, Russia}

\ead{$^*$Corresponding author e-mail:  klikunova@volsu.ru}

\begin{abstract}\footnotetext{\underline{Cite as}: Klikunova A.Yu., Khoperskov A.V. Creation of digital elevation models for river floodplains // CEUR Workshop Proceedings, 2019, vol. 2391,  275-284}
A procedure for constructing a digital elevation model (DEM) of the northern part of the Volga-Akhtuba interfluve is described.
The basis of our DEM is the elevation matrix of Shuttle Radar Topography Mission (SRTM)  for which we carried out the refinement and updating of spatial data using satellite imagery, GPS data, depth measurements of the River Volga  and River Akhtuba stream beds.
The most important source of high-altitude data for the Volga-Akhtuba floodplain (VAF) can be the results of observations of the coastlines dynamics of small reservoirs (lakes, eriks, small channels) arising in the process of spring flooding and disappearing during low-flow periods.
A set of digitized coastlines at different times of flooding can significantly improve the quality of the DEM.
The method of constructing a digital elevation model includes an iterative procedure that uses the results of morphostructural analysis of the DEM and the numerical hydrodynamic simulations of the VAF flooding based on the shallow water model. 
 
\end{abstract}

\section{Introduction}
 A high-resolution 3D topographic model for the large areas is essential to solving a variety of applied problems in the geosciences  that are associated with modeling and monitoring the environment.
 The progress of computer technology and numerical methods gives us new opportunities for modeling fluid dynamics in certain territories.
  Such problems include storm surges, spring floods in river valleys, flooding due to heavy rainfall \cite{LitKlikunova-Khrapov2013,Madadi2015}.
Hydrodynamic models allow technical and environmental expertise in the design of hydrological structures \cite{LitKlikunova-Agafonnikova2017,Jafary2018DSSwater}. 
 The important tasks are the determination of the watersheds' boundaries \cite{LitKlikunovaMaltcev2014}, the creating tools to help authorities respond to emergency situations \cite{Efremova2018mappingFlooded}.  
 
One important research area is the creation of decision support systems (DSS) for solving various hydrological problems, and the effectiveness of these DSS is determined by the quality of the applied digital elevation models (DEM) \cite{Voronin2017, Boori2018hyperspectral, Keenan2019SDSS}.
 Such DSS belong to the class of Spatial Decision-Support System, which combine standard decision-making tools with geographic information systems, providing new opportunities for water resources management \cite{Jafary2018DSSwater, Jonoski2016DSS}, city and regional planning, real-time decision-making for land management \cite{Grima2017DSS}, transportation engineering \cite{Bagloee2017DSStransport}, protecting the natural resources in conditions of increasing human pressures on the ecosystem.
 
A quality DEM is a critical component for all these tasks \cite{LitKlikunovaSatge2015}. 
The terrain is a major physical factor that influences the dynamics of water. 
Unfortunately, the accuracy of best topographic maps is not high enough for numerical simulations. 
In addition, new problems appear on small spatial scales, and they are associated with changes in the surface of the relief caused by natural and man-made factors
\cite{Ahmed2014,Reichenbach2018}. 
 Changes in the profile of the bottom and adjacent areas are a continuous process due to active sediment transfer and erosion processes, which require the use of the self-consistent model of water and sediment dynamics and regular updating of the DEM also~\cite{SidoryakinaSukhinov2017}.
 
In this paper, we describe the key stages of creating a DEM for river systems based on the synthesis of various spatial data using the example of the northern part of the Volga-Akhtuba floodplain (VAF). 
 The Volga Hydroelectric Station controls the flow of water downstream of the Volga River and the moisture reserves for the entire floodplain. 
 The volume flow of water through the dam is called discharge $Q(t)$ (m$^3\cdot$sec$^{-1}$) and it varies between $Q(t) \simeq 4000 - 30000 $\,m$^3\cdot $sec$^{-1}$ during the year.

\begin{figure}[t]
	\centering{
		\includegraphics[width=0.95\textwidth]{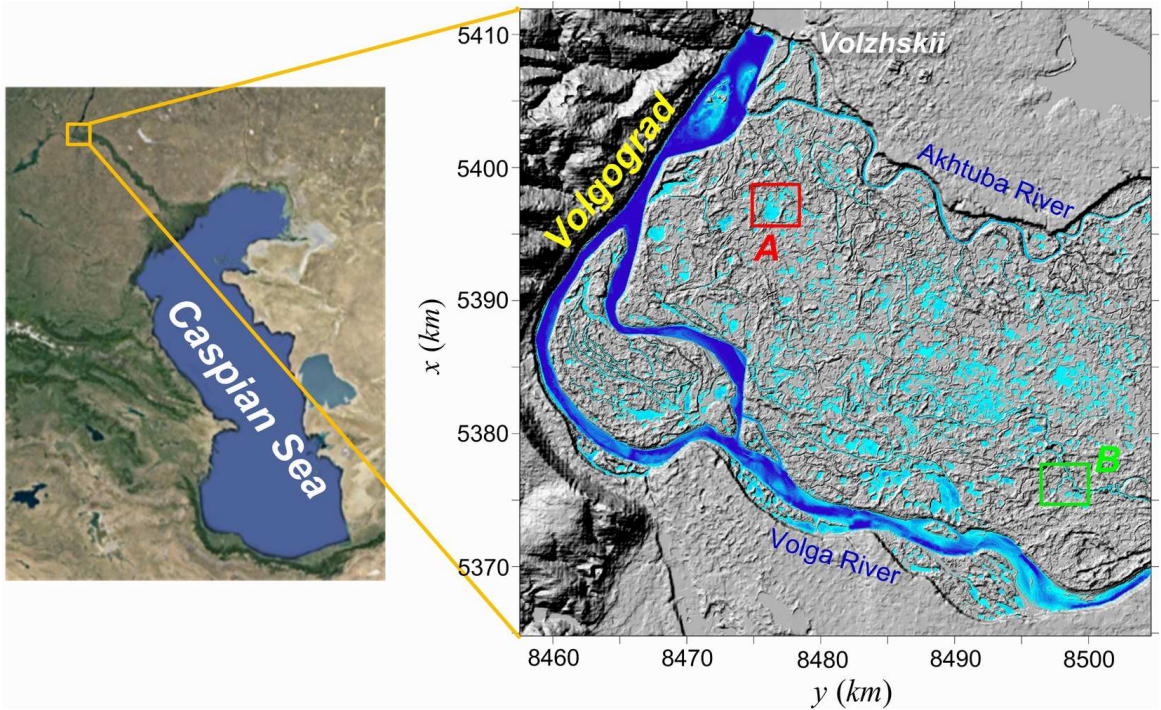}
}
	\caption{The northern part of the Volga-Ahtuba floodplain.}
		 	\label{FigKKlikunova01}
\end{figure}
 
 Important components of our methodology are the use of observational data on the dynamics of the coastlines of numerous small reservoirs  in the interfluve during the spring flood and the verification of DTM using hydrodynamic modeling.
Observations of the coastlines motions for a large number of reservoirs during the spring flooding are a source of very accurate local topography data.
 These water reservoirs are the results of the passage of spring water and they usually disappear in early summer.
Thus, the water surface area in the territory of VAF varies strongly during a few weeks from 2-5\% before flooding (low water) up to a maximum value of 20-40\%, which depends on the specific conditions in each year. 
  In late summer, the water basin area is smaller than in the early spring period before the flood, that connected with high summer temperature and lack of rain.
  The coastline coincides with the contour line (isoline) of the heights' distribution with very high accuracy at each time point.
  Thus, the local DEM may be the result of processing the monitoring data of the coastlines dynamics for a large number of small reservoirs during the spring flood.
These local DEMs are high-resolution data for the most critical areas in terms of hydrology as a part of global DEM for the northern territory of the Volga-Akhtuba floodplain (Fig.~\ref{FigKKlikunova01}).

\section{Iterative process of creating DEM}

\begin{figure}[t]
	\centering{
		\includegraphics[width=0.99\textwidth]{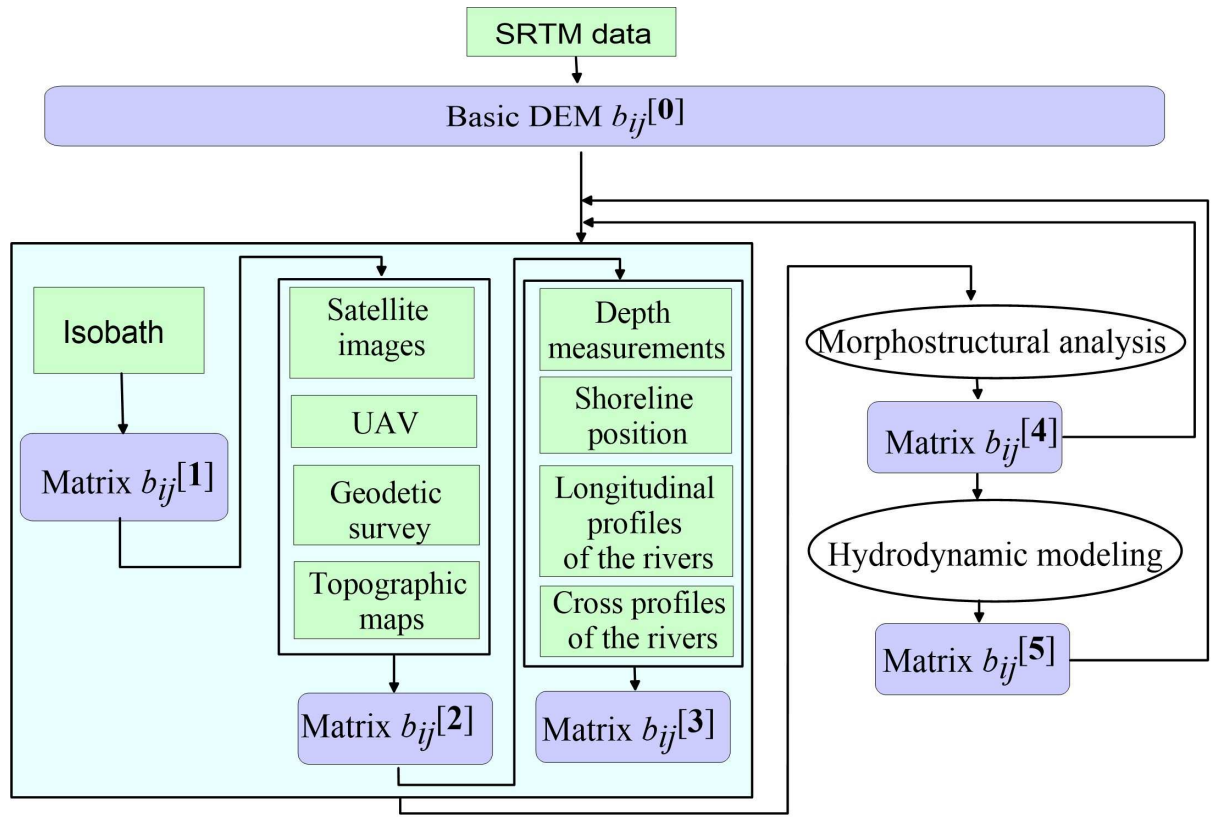}
	}
	\caption{\label{FigKKlikunova02}
		Stages and sequence of DEM creation. 	
	}
\end{figure}

\subsection{Main stages of creating DEM}
Figure~\ref{FigKKlikunova02} shows the general scheme for constructing a digital elevation model and highlights the most significant steps which we will discuss below.
Our DEM is based on the height matrix $b_{ij} = b(x_i, y_j)$ for nodes of the Pulkovo 95 coordinate system with step $\Delta {x} = \Delta{y}$:
 $x_i=x_0+i\Delta{x}$, $y_j=y_0+j\Delta{y}$ ($i=1,2, ..., N_x$, $j=1,2, ..., N_y$).
 We take the SRTM3 SRTMGL1 data $b_{ij}^{[SRTM]}$  as the initial height matrix.
 The professional GIS ``Panorama'' tools allow us to recalculate the matrix by a smaller step ($\Delta{x}=15$\,m, 10\,m, 5\,m) using the weighted average interpolation in 16 directions.
 Such matrix $b_{ij}^{[0]}$ will be called the basic digital elevation model.

The main stages of the transformation matrix $b_{ij}^{[0]}$ are discussed below.
 
 1) To clarify the model of the bottom of the Volga River and the Akhtuba River, we use \textbf{Sailing Directions (shipping charts) and water depth maps}.
 To refine the bottom model of the Volga River and the Akhtuba River, we use Sailing Directions (shipping charts) and reservoir depth maps, and then we obtain the matrix $b_{ij}^{[1]}$ after digitizing and embedding this data into the basic DEM $b_{ij}^{[0]}$.

\begin{figure}[h]
	\begin{center}
		\includegraphics[width=0.65\linewidth]{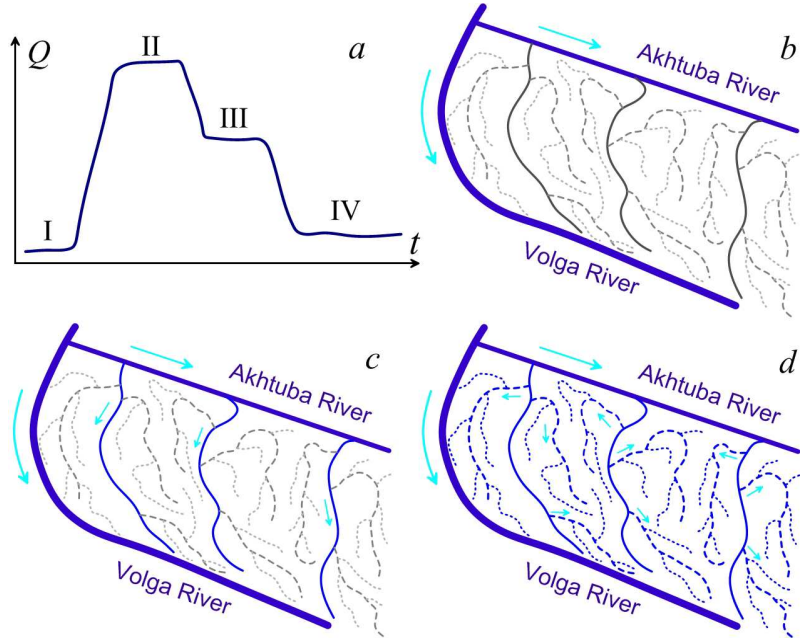}
		\caption{ \textit{a} --- Typical dependence of discharge $Q(t)$.
			\textit{b, c, d} --- The hierarchical structure of the hydrological system in the VAF at different stages of flooding.}\label{Fig-HydroRegime}
	\end{center}
\end{figure}

2) A unique feature of the VAF is a complex system of small channels in the interfluve (the so-called eriks), which form a hierarchical system of channels between River Akhtuba and River Volga (Fig.~\ref{Fig-HydroRegime}). 
  We use the \textbf{satellite images} of the ``RESURS-P'' series and UK-DMC\,2, the DigitalGlobe's satellite constellation (Google Earth services) to vectorize the linear objects of this channel system for subsequent introduction into the DEM matrix of $b_{ij}^{[1]}$.
UAV images and geodesic data are an important source for clarifying the location of small channels (Fig.~\ref{fig-klikunova-UAV}). 
As a result, we have the matrix $b_{ij}^{[2]}$, which contains the system of small channels.

\begin{figure}[h]
	\centering{
		\includegraphics[width=0.8\linewidth]{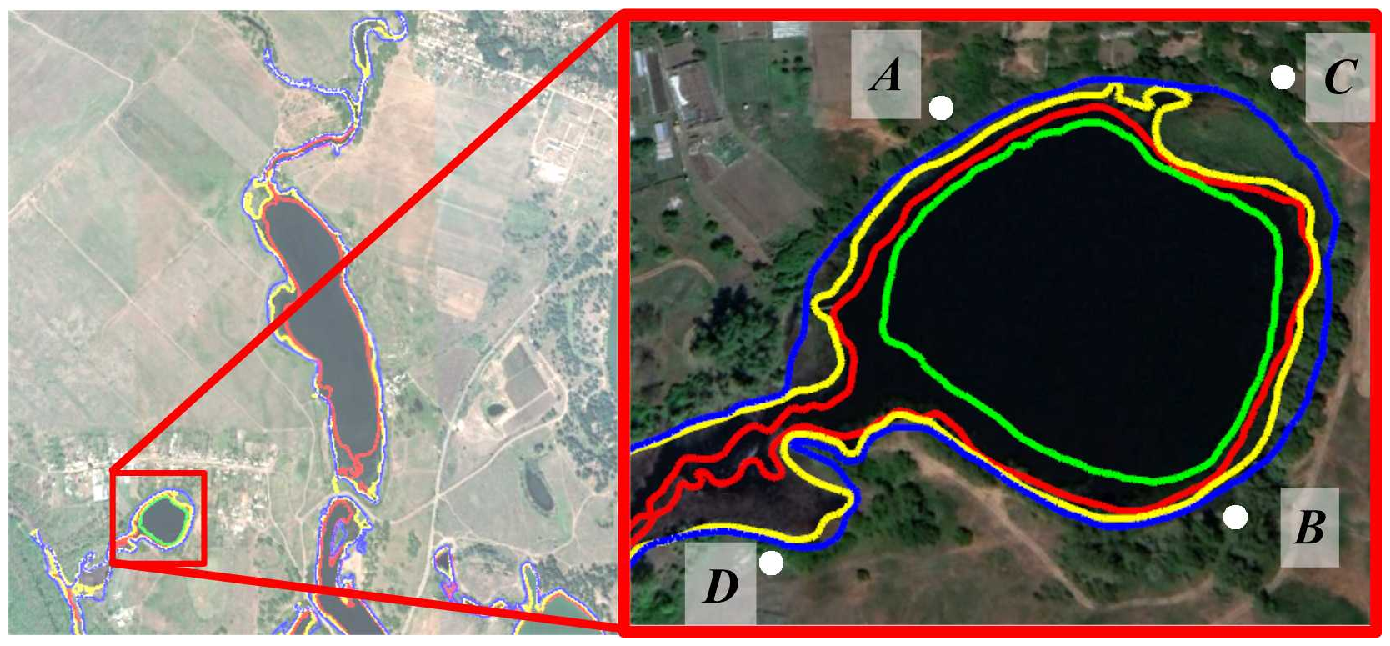}\\ a)\\
	    \includegraphics[width=0.8\linewidth]{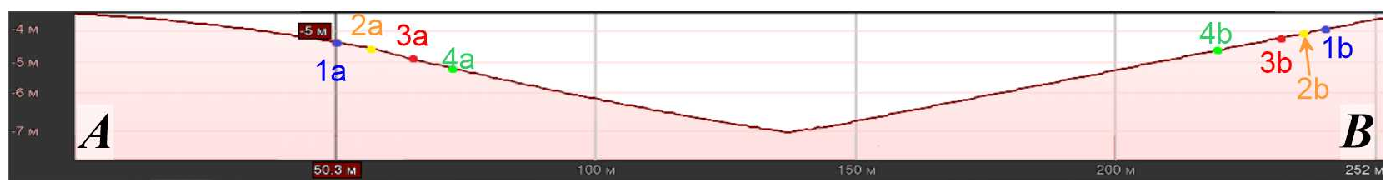}\\ b)\\
		\includegraphics[width=0.8\linewidth]{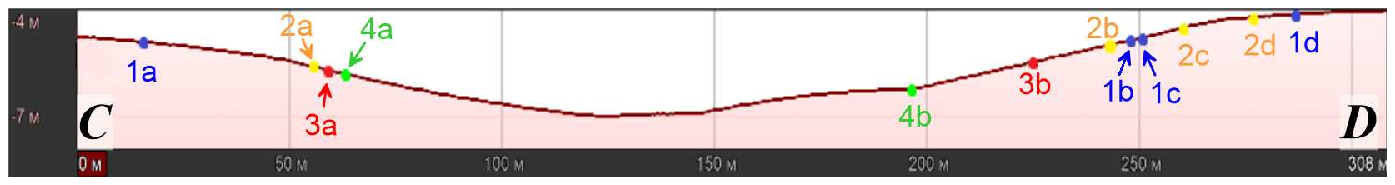}\\ c)\\
 }
	\caption{The position of coastlines at different points in time for small bodies of water near the village Zonal'nyj. \label{fig-klikunova-Zonal}
	 }
\end{figure}

3) To update the Volga River bottom model, we use the data of the last \textbf{depth measurements} ranging from the Volga hydroelectric power station to the Svetly Yar settlement.
These data are very sparse and after approximation to all our grid nodes  we have the matrix $b_{ij}^{[3]}$ with the height data of the river bed.

4) We use data on \textbf{dynamics of coastlines} of transient reservoirs, which are filled with water at the stage of interfluve flooding (April -- May) and dry out in the summer (Figure~\ref{fig-klikunova-Zonal}).
 These measurements provide an additional set of lines with a constant level of relief with very high accuracy.
 The refined matrix $b_{ij}^{[3]}$ is the result of binding these isolines to heights.
 Our studies have shown the effectiveness of the UAVs use to obtain data on the boundaries of water bodies (Fig.~\ref{fig-klikunova-UAV}).
 UAVs provide a more detailed sequence of isolines at the initial stage of flooding rise,  which is almost unattainable for satellite data. 
 However, this approach is local and does not allow to cover large areas.

\begin{figure}[h]
	\begin{center}
		\includegraphics[width=0.7\linewidth]{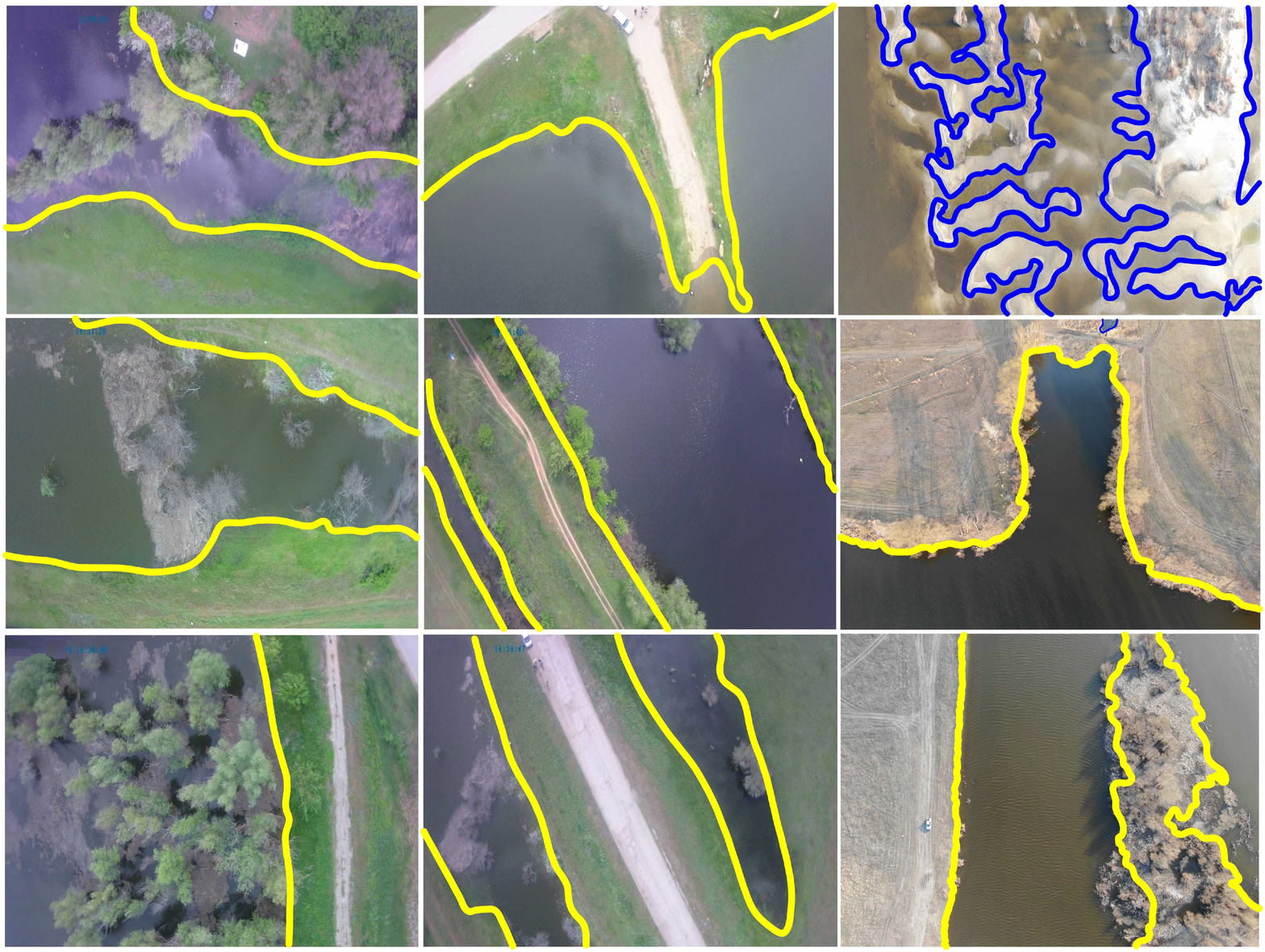}
		\caption{The vectorization of water bodies images with UAV. The colored lines show the boundaries of the reservoirs. \label{fig-klikunova-UAV}}
	\end{center}
\end{figure}

 Figure~\ref{fig-klikunova-Zonal} shows vertical profiles along the $AB$ and $CD$ segments for the $b_{ij}^{[2]}$ matrix, indicating the positions of the corresponding intersections of coastlines with these segments. 
 The points for the same coastline on opposite slopes of the reservoir have different elevation levels, which indicates the need to update the matrix $b_{ij}^{[2]}$.
 For example, the height difference is $\Delta b = 0.5$\,m for a pair of points (1a, 1b) in the figure\,\ref{fig-klikunova-Zonal}\,b and $\Delta b = 1$\,m for (2a, 2d) in the figure\,\ref{fig-klikunova-Zonal}\,c.

\begin{figure}[h]
	\begin{center}
		\includegraphics[width=0.9\linewidth]{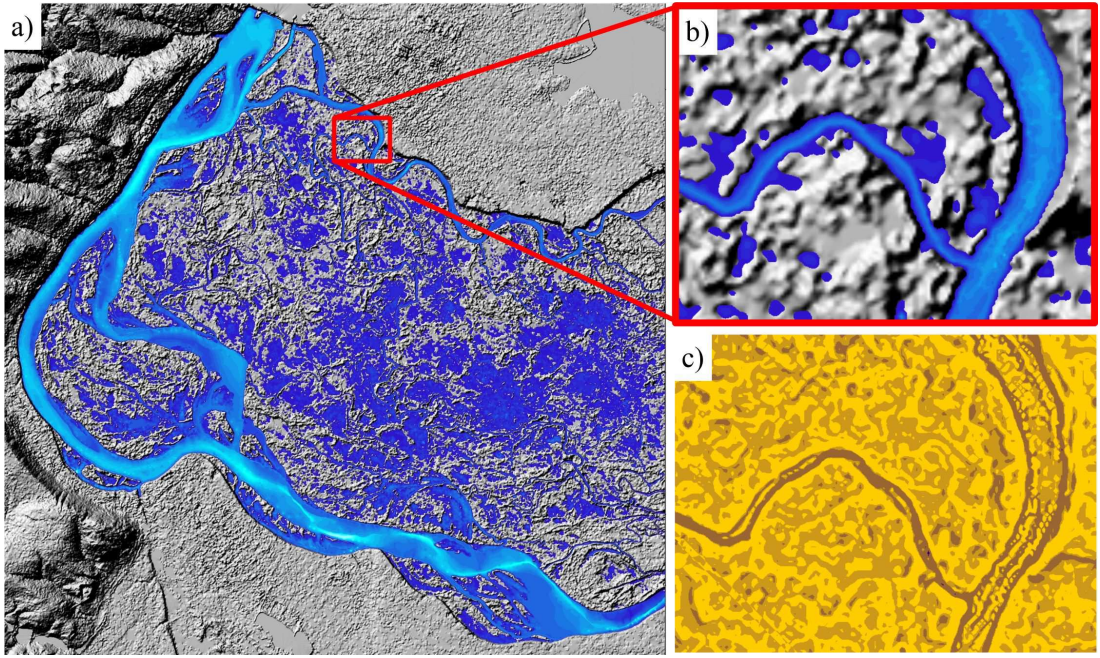}
		\caption{\textit{a}) The general structure of the VAF flooding is based on the results of our numerical hydrodynamic modeling. \textit{b}) The distribution of water for the specified area of the frame. \textit{c})~The distribution of the morphometric index $k_s$ for the same zone. \label{fig-klikunova-morpho}}
	\end{center}
\end{figure}

5) Then we calculate the standard set of \textbf{morphostructural analysis} parameters \cite{LitKlikunova-Sorokhtin2018}: the profile curvature  $k_t(x_i, y_j)$, the tangential curvature $k_s(x_i, y_j)$ and the tilt angles $s(x_i, y_j)$ (Figure\,\ref{fig-klikunova-morpho}):
\begin{eqnarray}
s&=&\frac{360^o}{2\pi}\arctan{\sqrt{b^2_x+b^2_y}}\,, \label{eq-s} \\ 
k_t&=&\frac{b_{xx}b^2_y-2b_{xy}b_x b_y+b_{yy} b^2_x}{p\sqrt{q}}\,, \label{eq-kt} \\
k_s&=&\frac{b_{xx} b^2_y+2b_{xy}b_x b_y+ b_{yy} b^2_x}{p\sqrt{q^3}}\,, \label{eq-ks}
\end{eqnarray}
где $\displaystyle b_x=\frac{\partial b}{\partial x}$, $\displaystyle b_y=\frac{\partial b}{\partial y}$, $\displaystyle b_{xx}=\frac{\partial^2 b}{\partial x^2}$, $\displaystyle b_{yy}=\frac{\partial^2 b}{\partial y^2}$, $\displaystyle b_{xy}=\frac{\partial^2 b}{\partial x \partial y}$, $\displaystyle p=b^2_x+b^2_y$, $q=1+p$. 
 
 We often encounter two types of artifacts:

\noindent
 a) Strong local errors of heights on the $b_{ij}^{[0]}$ matrix are strongly highlighted against the background of a rather flat territory.
 These errors are often caused by data processing problems for small forests and small water reservoirs.

\noindent
 b) The second difficulty is related to the detection of small channels connectedness.

 There are problems with the automatic selection of objects even in images for urbanized areas, the morphology of which is simpler compared to the wooded marsh landscape of the floodplain \cite{Michaelsen2018compOpt}.
 Analysis of the hyperspectral observational data for various platforms allows us to improve the classification of objects \cite{Boori2018hyperspectral}, but this approach is algorithmically complex \cite{Myasnikov2017Hyperspectral}.
 The spatial distributions of the parameters (\ref{eq-s})\,--\,(\ref{eq-ks}) help identify areas with artifacts, first of all, areas with a violation of hydrological connectedness of watercourses on the digital elevation model.
 The morphostructural analysis of the DEM allows simple means to detect possible errors and promptly correct them, refining the hydrological network \cite{LitKlikunovaRueda2013, LitKlikunovaElmahdy2015}. 

6) Hydrodynamic modeling is carried out at the final stage (Fig.~3\textit{a}, \textit{b}), reproducing the spring flooding of the interfluve territory in accordance with the procedure described in \cite{LitKlikunova-Khrapov2013, LitKlikunova-Agafonnikova2017, LitKlikunova-Khoperskov2018}.
 This allows you to check the channels connectedness of the hydrological system in addition to the morphostructural analysis.
 Comparison of simulation results with observational data is a powerful tool for updating the DEM for the most important zones, which primarily provide for the formation of vast reservoirs of the lake type due to the water outflow from small canals (eriks). 
 Such verification based on hydrodynamic modeling is the most resource-intensive procedure.
 For hydrodynamic simulations, we use the software for the numerical solution of the shallow water equations described in \cite{LitKlikunova-Khrapov2013, LitKlikunova-Khoperskov2018} and taking into account the parallel implementation for GPUs \cite{LitKlikunovaDyakonova2016}.

\subsection{Assimilation of local spatial data by the DEM matrix}
 One essential feature of building a digital model of river bed is the  source data sparseness, which include:
\begin{enumerate}
	\item There are two coastlines with water level mark $L_{1}^{coast}(\vec{r})$, $L_{2}^{coast(\vec{r})}$.
	\item There are several depth curves on topographic maps of $L_{i}^{bed}(\vec{r})$ ($i = 1, ... , m_B$).
	We have only $m_B\sim 3-4$ even for the largest rivers.
	\item Several soundings show, as a rule, only the deepest points on a topographic map.
	\item 
	Depth measurements using echo sounders require new field studies.	
\end{enumerate}
All these data form set of points ${\cal P}$ on the height matrix $b_{ij}$.

\begin{figure}[h]
		\begin{minipage}[h]{0.5\linewidth}
\center{\includegraphics[width=1.05\linewidth]{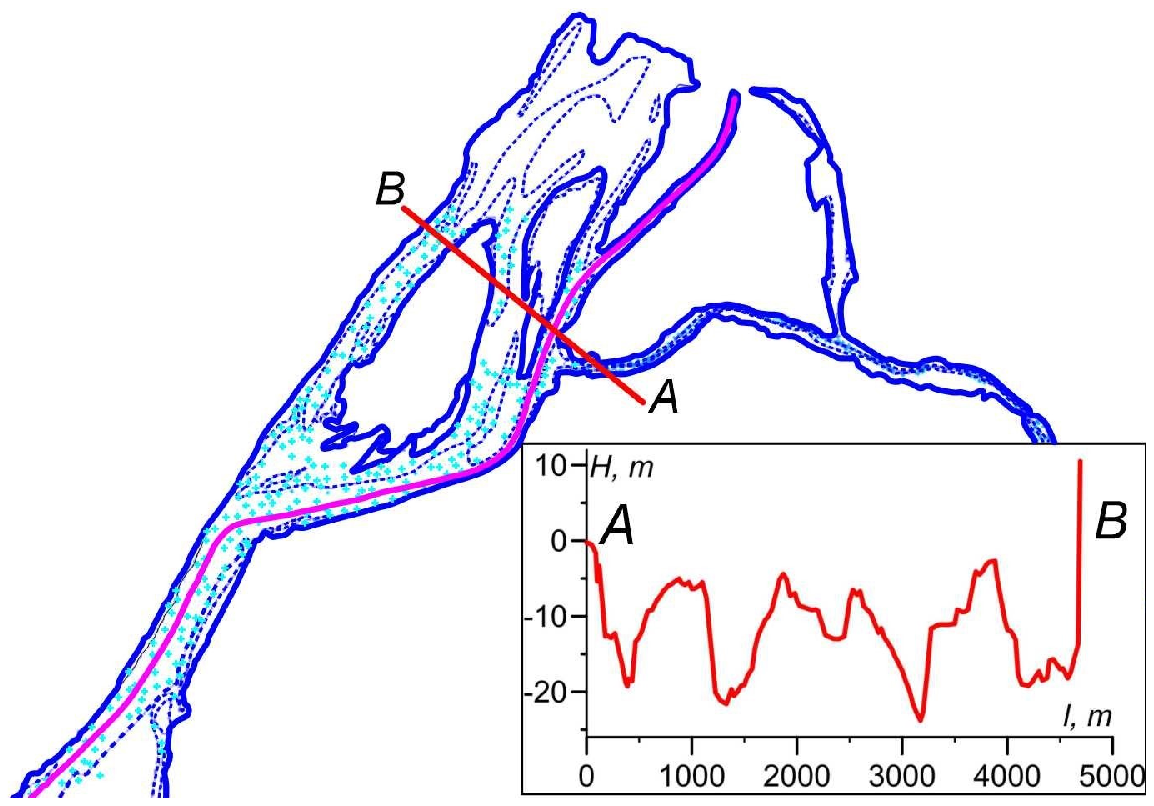} \\ a)}
\end{minipage}
\hfill
\begin{minipage}[h]{0.45\linewidth}
\center{\includegraphics[width=1\linewidth]{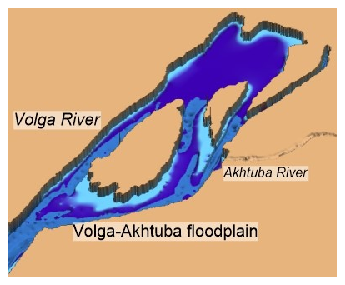} \\ b)}
\end{minipage}
\caption{\textit{a}) Vector map of the River Volga. 
	\textit{b}) Digital elevation model of riverbed of the Volga downstream from the hydroelectric dam.}
\label{ris:image1}
\end{figure}

We used an iterative procedure to build a river bottom DEM:
\begin{equation}\label{eq-Poisson}
 b_{n,m}^{p+1} = \left\{ \begin{array}{lc}
    b_{n,m}^{p}+\alpha \left[ b_{n+1,m}^{p}-2b_{n,m}^{p}+b_{n-1,m}^{p} \right]+\alpha\left[ b_{n,m+1}^{p}-2b_{n,m}^{p}+b_{n,m-1}^{p} \right]\,, & P_{n,m} \notin {\cal P} \\
  b_{n,m}^{(exp)} \,, & P_{n,m} \in {\cal P}\,,    \\
  \end{array} \right.
\end{equation}
where $b_{n, m}^{(exp)}$ is the depth at the points ${\cal P}_{n, m}$, $\alpha$ is the parameter that determines the convergence of the iterative procedure (\ref{eq-Poisson}).
The formula (\ref{eq-Poisson}) is the finite-difference analog of the diffusion equation.
 We obtain the solution to the Poisson's equation in the case of converging iterations~(\ref{eq-Poisson}).
 Figure~\ref{ris:image1} shows the results of the construction of the DEM of the Volga River area, based on the approach described above.

\subsection{Coastlines dynamics as factor in improving DEM}
Fig.~\ref{Fig-HydroRegime} shows a schematic diagram of the hydrological regime in the VAF. 
Water flows from the Volga River to the Akhtuba River in a low water period in the case $Q\simeq 5-9$ thousands m$^3\cdot$sec$^{-1}$, but it is not enough to fill the channels and besides the moisture reserve is very small in the area between the rivers. 
All channels are quickly filled with the increase of $Q$ up to 23--30 thousands m$^3\cdot$sec$^{-1}$  and the water is poured onto the flat part of VAF.
  The water level is maintained by the powerful moistening at the third stage with $Q = 16000-20000$\,m$^3\cdot$sec$^{-1}$. 
  In late spring, there is a change to low-water and the total moisture content decreases in the territory. 

\begin{figure}[h]
	\begin{center}
		\includegraphics[width=0.70\linewidth]{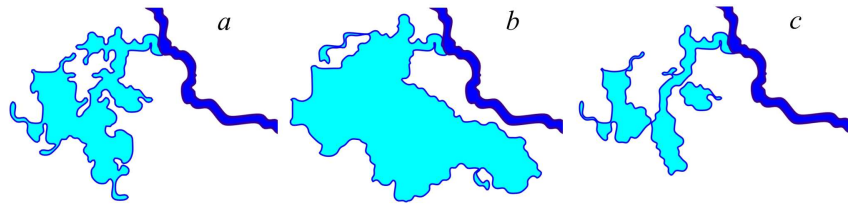}
		\caption{Shallow lake near the Bulgakov Channel at various stagesof the flooding in 2014 is shown: a) start the flooding (May 6), b) maximum the flooding (May 8), c) the dissipation of the reservoir (May 18).}\label{Fig-bulgakov}
	\end{center}
\end{figure}

There is a large number of shallow lakes on the flat territory between the large and small channels in spring and early summer. 
The coastlines of such reservoirs are moved on considerable distances in a short time period (Fig.~\ref{Fig-bulgakov} and See Fig. \ref{fig-klikunova-Zonal}). 
 Measuring the position of coastline at different points in time can help us determine an additional set of contour lines (isolines of heights) of the terrain for critical zones.

\subsection{Verification based on the results of hydrodynamic simulations}

\begin{figure}[h]
	\begin{minipage}[h]{0.47\linewidth}
\center{\includegraphics[width=1\linewidth]{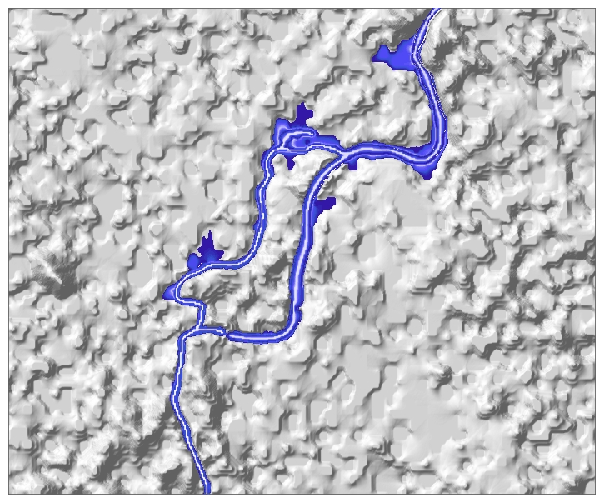}} a) \\
\end{minipage}
\hfill
\begin{minipage}[h]{0.47\linewidth}
\center{\includegraphics[width=1\linewidth]{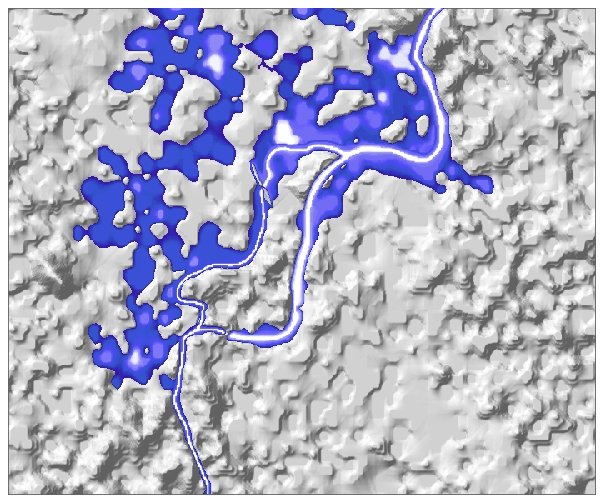}} \\b)
\end{minipage}
\vfill
\begin{minipage}[h]{0.47\linewidth}
\center{\includegraphics[width=1\linewidth]{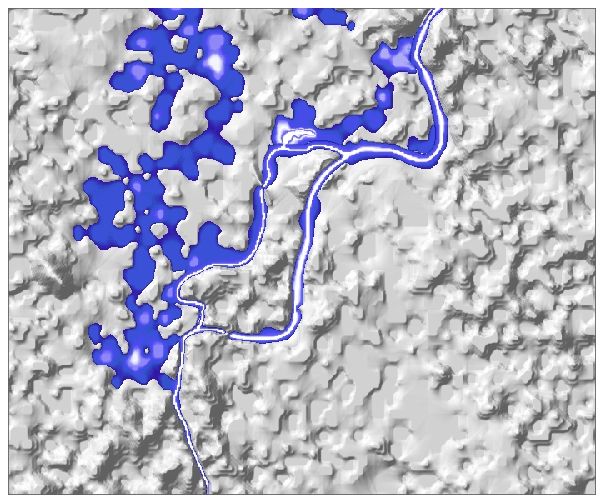}} c) \\
\end{minipage}
\hfill
\begin{minipage}[h]{0.47\linewidth}
\center{\includegraphics[width=1\linewidth]{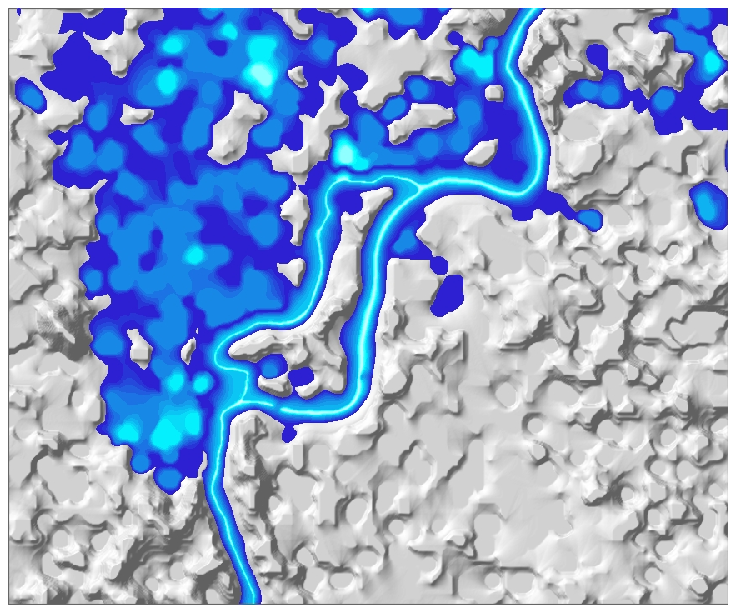}} d) \\
\end{minipage}
\caption{Results of local DEM refinement for the small river valley using hydrodynamic simulations. 
	By identifying the shortcomings of the DEM, we provide flooding in the model for the nearest areas in accordance with the observations.
 }
\label{fig-hydroSimul}
\end{figure}

Fig.~\ref{fig-hydroSimul} shows the results of hydrodynamic simulations in the floodplain of the small river at various stages of the DEM refinement:
\begin{enumerate}
	\item We use the DEM after embedding the riverbed in the SRTM matrix and assignment the coastlines, the fairway line and the river slope (Fig.~\ref{fig-hydroSimul}\textit{a}).
	\item Panel \textit{b} in the figure demonstrates the water distribution in the river channel after processing the DEM in the ``Construction of horizontals by elevation matrix'' service in the GIS Panorama.
	\item The next iteration involves the DEM rebuilding taking into account the geodetic transverse profiles of the river valley, which are obtained as a result of field measurements (Fig.~\ref{fig-hydroSimul}\textit{c}).
	\item The final step involves updating the digital model on a small scale at the high water stage (Fig.~\ref{fig-hydroSimul}\textit{d}).
\end{enumerate}

\section{Conclusions}

The object of our study is the valley between the River Volga and River Akhtuba, the ecosystem of which is unique on Earth due to the special hydrological regime.
 We propose the iterative procedure for creating the DEM for special floodplain areas with a large number of transient reservoirs.
 The initial data are the SRTM matrix, the space images from the ``Resource-P '' and UK-DMC-2 satellites, the topographic maps, the geodetic measurements of the elevation profiles, the depth measurements.
 The morphostructural analysis and the numerical simulations of surface water dynamics on realistic topography can be powerful tools for verification of the digital elevation model.

 The observed dynamics of coastlines allows building elevation levels along the boundaries of water bodies, and this approach is actively used to construct the DEM.
 However, this method acquires special value in the case of periodically flooded areas, since the moving coastlines provide detailed sets of contour lines, being the basis for a very high-quality and relevant digital elevation model.

\ack{The work has been supported by the Ministry of Science and Higher Education  (government task no. 2.852.2017/4.6). The research is carried out using the equipment of the shared research facilities of HPC computing resources at Lomonosov Moscow State University.
	The authors are grateful to E.~Agafonnikova, S.~Khrapov, A.~Pisarev, K.~Tertychny for their help and assistance in carrying out this project. 
	A.~Klikunova thanks for the support of the RFBR and the Administration of the Volgograd region (grant 18-47-340003). 
}

\end{document}